\documentclass[letterpaper, 10 pt, conference]{ieeeconf}  
\IEEEoverridecommandlockouts
\usepackage[utf8]{inputenc}
\usepackage{authblk}
\usepackage{graphicx}
\usepackage{fixltx2e}
\usepackage{lipsum}
\graphicspath{ {./Pix/} }
\title{\LARGE \bf
Deep Learning in Healthcare: An In-Depth Analysis
}
\author[1]{Farzan Shenavarmasouleh}
\author[1]{Farid Ghareh Mohammadi}
\author[1]{Khaled M. Rasheed}
\author[1]{Hamid R. Arabnia}
\affil[1]{Department of Computer Science, University of Georgia, Athens, Georgia, United States \authorcr {\{fs04199, farid.ghm, khaled, hra\}@uga.edu}\vspace{1.5ex}}
\begin{document}
\maketitle
\thispagestyle{empty}
\pagestyle{empty}
\begin{abstract}
Deep learning (DL) along with never-ending advancements in computational processing and cloud technologies have bestowed us powerful analyzing tools and techniques in the past decade and enabled us to use and apply them in various fields of study. Health informatics is not an exception, and conversely, is the discipline that generates the most amount of data in today's era and can benefit from DL the most. Extracting features and finding complex patterns from a huge amount of raw data and transforming them into knowledge is a challenging task. Besides, various DL architectures have been proposed by researchers throughout the years to tackle different problems. In this paper, we provide a review of DL models and their broad application in bioinformatics and healthcare categorized by their architecture. In addition, we also go over some of the key challenges that still exist and can show up while conducting DL research.

\vspace{\baselineskip}

\end{abstract}
\section{\textbf{INTRODUCTION}} 
Healthcare industry generates almost a third of all the data on the planet annually \cite{rydning2018digitization}. Every second, a massive amount of data is generated from all types of devices, imaging tools, patient management portals located within the hospitals and clinics, or from all the emerging Internet of Things (IoT) sensors and implants. This huge load of heterogeneous data that contains highly redundant and correlated info is called Big Data \cite{sagiroglu2013big}. Extracting valuable knowledge from big data is inherently a difficult task. But, analyzing healthcare data enables us to take modern diagnosis and treatment to an entirely new level. It can aid doctors and clinicians in analyzing radiology images with more accuracy and direct their attention to the most important locations of the picture and thus help them find the nodules, masses, and other deficiencies that otherwise would have been missed. Aside from this, they can work as stand-alone classifiers and/or clustering tools to diagnose various diseases and even predict the malignity of them. Additionally, they can be utilized in creating novel personalized medicines and help researchers in genomics.

Over the past few decades, Machine Learning proved itself as a valuable analysis technique and researchers have benefited from all the different powerful methods that come with it, such as support vector machine (SVM), tensor decomposition, random forests, Bayesian networks, and much more, to extract useful features and discover hidden patterns within the data. However, for the most part, using these methods required engineering custom features by experts with extensive knowledge about the domain of the task at hand and this led to the emergence of a new branch of machine learning, so-called deep learning that occupied researchers with much more sophisticated tools and methods that overcame the previous shortcomings. A Deep Learning architecture can be viewed as an Artificial Neural Network (ANN) with two or more hidden layers and it is capable of extracting high-level features from data automatically and using that to perform the task, thus removing the need for expensive and time-consuming feature engineering phase while yielding a better accuracy.

With the advancement of deep learning, it is getting employed in more and more fields by researchers and it is now a must-have in most of the interdisciplinary studies. Healthcare and bioinformatics are not exceptions. A wide range of deep learning architectures has been used to analyze the massive data in healthcare throughout the years. In this article, we aim to provide a review of deep learning methods, their diverse applications for bioinformatics and healthcare research categorized by their most prominent architecture, along with the challenges that come with them. We believe that this article can serve as a starting point for researchers and provide valuable insight for future studies in deep learning in bioinformatics research.


\section{\textbf{Classification and Segmentation Models}}

\subsection{\textbf{Convolutional Neural Networks}} \label{cnn} 
The ultimate objective of computer vision is to mimic the behavior and operation of human eyes. Convolutional Neural Network (CNN) is a machine learning algorithm developed based on biological visual cortex processing and it aids computers in analyzing pictures and movies and, as a result, understanding the objects contained in them. CNNs have become the go-to technique for any work that requires dealing with media after AlexNet \cite{krizhevsky2012imagenet} won the ImageNet image classification competition in 2012. 

CNN is a supervised deep learning architecture that employs three types of layers namely, convolutional layers, pooling layers, and fully connected layers to solve all sorts of different classes of tasks. The top five categories are Image Classification, Object Localization, Object Detection, Semantic Segmentation, and Instance Segmentation, listed in ascending order of complexity. 

In an Image Classification task, usually, a single core object is present in the image, and the aim is to determine which category that image belongs to. Object Localization is a little more complex. The model's aim in object localization is to output the position of those items as bounding boxes, a rectangular box surrounding the object, in addition to predicting the category to which they all belong to.

Object detection comes next. CNNs are capable of detecting edges and, as a result, determining object boundaries. As a consequence of this characteristic, they can be used to detect a variety of objects in a given image. However, doing so necessitates applying them to a large bunch of areas with varying scales on each image, which takes a long time. As a result, a significant amount of study has been conducted to address this problem.

It all started with R-CNN (Region-based CNN) in \cite{girshick2014rich} where the authors used a region proposal module to tackle the aforementioned problem. Following this publication, the same authors went deeper and improved the model by addressing the model's overlapping nature in previous work, eliminating unnecessary calculations, and allowing the framework to train all three concurrent models at the same time. This model was given the name Fast R-CNN \cite{girshick2015fast}. Finally, Faster R-CNN \cite{ren2015faster} was proposed to fix the bottleneck that existed within the Fast R-CNN by employing Region Proposal Network (RPN), altogether making Faster R-CNN the go-to model for most of the use cases regarding object detection.

Object detection, as great as it is, is still unable to comprehend and supply us with the actual shape of the objects, and instead just provides the bounding boxes. This is where Image Segmentation shines since it solves the problem by producing a pixel-by-pixel mask for each object. The task of image segmentation can be classified into two broad categories. Every pixel in the picture must be assigned to a predetermined class in Semantic Segmentation. Furthermore, all pixels relating to a class are treated the same and are given the same color, thus differences between various object instances belonging to the same class are ignored. Instance Segmentation, on the other hand, treats each instance of a certain class as a separate entity with its own color and label.

In the case of Semantic Segmentation, deep learning solutions started with \cite{long2015fully} where the authors presented a fully convolutional end-to-end trainable network. 
The big picture for architecture was for the model to have a well-known classification model such as AlexNet as encoder and then use transpose convolutional layers with pixel-wise cross-entropy loss as the decoder to upsample the result into the same size as the original image. However, since the encoder reduces the resolution of the image, the decoder failed to produce accurate masks. Thus, to counter this, the authors decided to add skip connections from earlier layers adding their values to later layers to offer required information for the decoder to properly create the masks more accurately. This approach has shown to be highly effective.
Following the success of this paper,  the authors of \cite{ronneberger2015u} presented the U-Net architecture which consists of a contracting path to capture context and a symmetric expanding path for accurate localization. U-Net gained a lot of popularity, particularly in the medical field in which we will talk more about in detail in later sections. The idea of skip connections attracted a lot of attention and caused researchers to further study and analyze them that led to the creation of architectures such as DenseNet \cite{jegou2017one}, SegNet \cite{badrinarayanan2017segnet}, and ResNet \cite{he2016deep}.

The Mask R-CNN model \cite{he2017mask} was proposed to help image instance segmentation. It has built on Faster R-CNN by specializing in producing pixel-by-pixel masks for each object in addition to identifying the bounding boxes and class labels. It ingeniously adds another Fully Convolutional Network (FCN) on top of RPN, technically
adding a new parallel branch to the Faster R-CNN model architecture that generates a binary mask for the object discovered in a given region. It's also worth mentioning that the authors had to make minor changes to the model to address a problem with location misalignment caused by its quantization behavior. Authors in \cite{shenavarmasouleh2020drdr, shenavarmasouleh2020drdr2, shenavarmasouleh2021drdrv3} utilize Mask R-CNN to localize, mask, and detect the type of lesions that appear in the eyes of diabetic patients.

CNNs are primarily designed for fixed-size 2D images; however, most of the tasks in medical profession use MRIs and CTs which are inherently 3D or 4D with varying sizes, and with objects being relatively very small and being positioned in arbitrary locations. 
The naive solution is to use 3D images themselves as they are. Convolutional layers have been reconfigured and extended to 3D kernels to create 3D convolutional networks and they proved to surpass other approaches \cite{payan2015predicting, maturana2015voxnet}. However, processing 3D images using 3D convolutional networks is computationally intensive.

There has been an extensive amount of research to enhance this architecture. In \cite{cciccek20163d}, authors expanded U-Net from 2D to 3D architecture. In another similar work \cite{milletari2016v}, V-Net, a 3D version of U-Net architecture was presented. And authors in \cite{zhu2018deeplung} used 3D Faster R-CNN to segment nodules. The other two commonly used techniques were multi-stream learning and 2.5D models. 

Multi-stream aims to look at the data with varying perspectives, angles, directions, scales, resolutions, and even a combination of different types of data over the same period of time. Most of the time, the proposed architecture train one channel for each of the variations of the data and then concatenate them all into one main channel before doing the image analysis task such as classification or segmentation. 
Multi-scale analysis, for instance, comes from the intuition that blurring pictures with Gaussian Blur vanishes details that are smaller than a certain resolution. The power of multi-scale image analysis comes from the ability to change the sharpness of the image dynamically and hence look at details with different levels of resolution \cite{ter2010multi}. For example in \cite{bauer2011multiscale} authors analyze and segment tumors in brain images using multi-scale approaches. \cite{madabhushi2011computer} employs multi-scale features to create prognostic classifiers for predicting treatment response and patient outcome. In \cite{salvi2018multi} a multi-scale segmentation model, namely MANA was presented for nuclei detection. Authors in \cite{kawahara2016multi} propose a novel multi-resolution CNN and use it to detect skin lesions. And, \cite{yu2015multi} utilize dilated convolution \cite{wei2018revisiting} to perform multi-scale semantic image segmentation. In multi-modality approaches, researchers make use of different types of imaging and screening techniques all at the same time. Doing so provides two main advantages. First, this prevents the model from overfitting, and second, it helps different streams of data collected over the same period of data complement each other and cover the shortcomings of the others in the event of one not being able to catch a particular kind of detail about the data \cite{guo2018medical, gong2016multi, asali2020deepmsrf}. Similarly in 2.5D architectures, the researchers try to mainly convert $k \times k \times k$ media into k orthogonal $k \times k$ 2D channels with possible extra ones for additional information. \cite{xing20192} proposes 2.5D architecture for semantic segmentation and \cite{kushibar2018automated, xue2020multi} use 2.5D models for analyzing brain images and segmenting stroke lesions.

\subsection{\textbf{Autoencoders}} 
Autoencoder is a type of unsupervised deep learning model architecture that aims to encode data efficiently and use the encoding to reconstruct the original input with minimum loss. It accomplishes this by mapping input to itself through an interconnected neural network. Feature reduction, extraction of latent feature representations, and inferring missing data are some of the applications that autoencoders are inherently good at given their architecture characteristic.

There exist several variations on the base model each tailored for a specific task. Denoising and stacked denoising autoencoders, as their name suggests, are perfect for removing noise from the input data and can be used as a preprocessing step in all sorts of medical tasks \cite{gondara2016medical, majumdar2015real, liu2018low}. Sparse autoencoders add a sparsity penalty to the loss function and even though they can have more hidden units than inputs, only a small group of the hidden units are allowed to be active at the same time forcing the model to compress related unique features to each other. Sparsity improves performance on classification tasks. Thus, they have been utilized in various classification tasks such as breast cancer nuclei detection \cite{xu2015stacked}, and diagnosis of Alzheimer’s \cite{bhatkoti2016early} and Parkinson’s disease \cite{li2019longitudinal}. Sparse autoencoders can be stacked over each other and form stacked autoencoders which are more powerful in capturing more complex features \cite{wang2019optimization, ozturk2020stacked, almakky2018stacked}. Also, any kind of autoencoder with more than one hidden layer is called a deep autoencoder, again giving the model a better ability to understand inter-related characteristics among the input features. As an example, authors in \cite{mao2018feature} use deep autoencoders to classify nodules in lung images and researchers in \cite{song2017hybrid} also employ them to detect different types of cells in bone marrow biopsy images.

Unlike traditional autoencoders that map the data to a single value in latent space, variational autoencoders tend to map input features into a probability distribution for each latent attribute. Given this characteristics of them, they can easily act as generative models as well by providing the possibility to interact with latent space probabilities. Generative models such as generative adversarial networks (GAN) \cite{goodfellow2020generative} and variational autoencoders can be used for image super-resolution where a bigger or more enhanced image is needed such as in endomicroscopy \cite{ravi2019adversarial}, single molecules images \cite{speiser2019teaching}, and in cases where low-dose CT (LDCT) scans have also been considered instead of normal-dose CT (NDCT) to reduce potential health risks \cite{ma2011low}.

\subsection{\textbf{Deep Belief Networks}} 
Deep Belief Networks (DBN) are a multi-layer neural network with both directed and undirected connections. Although at the first glance, it appears to have a basic neural network design, the training procedure is vastly different between the two. Deep Belief Networks are a stack of Restricted Boltzman Machines (RBMs) that have been trained greedily layer by layer and fine-tuned using the up-down algorithm to learn high-dimensional manifolds of data as opposed to end-to-end training of neural networks using backpropagation. In neural networks, the layers are built on each other and each layer uses previous embeddings to create a more abstract meaning out of the data and learn higher-level features. In DBNs, however, each layer learns and encodes the entire input. Given the characteristics of DBNs, they can be incorporated in either supervised tasks as stand-alone classifiers or be used in unsupervised manners like for instance the way they are being employed in autoencoders or as generative models. 
DBN was employed in \cite{pinaya2016using} as a classifier to differentiate between healthy and schizophreniac patients, and it was shown that it outperforms a famous machine learning technique known as support vector machine (SVM). In \cite{khatami2017medical}, authors utilize DBN in an unsupervised manner to create a framework for analyzing radiology images. And in \cite{li2018latent}, authors use unsupervised DBN to extract more useful latent space features for fMRI images.

\subsection{\textbf{Recurrent Neural Network}} 
Recurrent Neural Network (RNN) is used for extracting patterns from sequential data and bringing in temporal features to the learning process. They can be utilized in all tasks that can benefit from this characteristic such as text processing and video processing. Since videos can be viewed as a sequence of images, in the context of biomedical imaging these models can be useful in any task that is related to the time such as analyzing the progression of diseases. RNN is the base model and in the past few years two more complex models, namely Long Short Term Memory (LSTM) \cite{hochreiter1997long} and Gated Recurrent Units (GRUs) \cite{cho2014learning}, have been emerged to address its shortcomings such as the vanishing gradient problem and by doing so extended the ability of sequential models and gave them the ability to keep track of older data more successfully. Later, their performance increased even more by incorporating attention techniques into recurrent models \cite{vaswani2017attention} as well.

In \cite{mao2019spatio}, authors employ LSTM to create a Spatio-temporal deep learning method that uses resting state functional magnetic resonance imaging (rs-fMRI) to diagnose Attention Deficit/Hyperactivity Disorder (ADHD). Authors in \cite{el2019hybrid} utilize same approach on fMRI images for autism disease classification. Spider U-Net \cite{lee2021spider} also utilizes LSTM to segment blood vessels in 3D using computed tomography and magnetic resonance angiography (MRA) images. In \cite{xu2019deep}, authors use GRU to predict lung cancer Treatment response in patients. \cite{gao2018fully} also proposed a deep learning method for 4D segmentation of longitudinal MRI while considering the brain maturation in infant brain imaging.

\subsection{\textbf{Reinforcement Learning}} 
Reinforcement learning (RL) is a machine learning approach that rewards desirable actions while penalizing undesirable ones. Doing this will force the agent to learn the best behavior, also known as policy, which leads to the most cumulative reward by utilizing Markov decision process (MDP) mathematical formulation. RL agents can constantly communicate with their environment via sensors and possible actuators and decide on their next action based on their current state. After enough trial and error, we end up with agents that know what they want in a given environment and can simply act on it to maximize their gain. 

Deep reinforcement learning extends RL and enables it to deal with higher-dimensional problems by bringing deep learning into the equation. This way, data types such as images that have a relatively large number of input size can benefit from be RL algorithms as they can be fed into deep RL models without worrying about having to manually engineer the state space of the problem and avoid the curse of dimensionality \cite{mohammadi2021evolutionary}.

RGB pixels of images cannot be used in RL to make decisions by themselves. Instead, CNNs, mostly 2D and 3D, are utilized as the primary feature extractors in the tasks that require working with media. These features can then be used to aid the model to decide on the optimal next move that enforces achieving the most expected return. In \cite{ghesu2017efficient} authors propose Marginal Space Deep Learning (MSDL) framework that uses RL to perform tasks of object localization and boundary estimation for arbitrarily shaped landmarks in medical images. This is particularly useful in finding tumors and cysts with varying sizes and shapes in images taken from the patients and successfully locating and masking them using RL methods. This is comparable to methods mentioned in section 2.1 that perform semantic and instance segmentation. Similarly, \cite{ghesu2017multi} uses deep RL to perform 3D-landmark detection in CT scans in real-time. \cite{navarro2020deep} make use of a deep RL model to localize organs in the body by running agents on CT images. And in \cite{liu2019deep} authors employ deep RL and Q-learning techniques for the task of lung cancer detection.


\section{\textbf{More Applications}} 
In previous section we went through the most prominent deep learning models, saw their progression through the years, and learned about a few of their use cases mostly in the medical industry. Aside from these models, deep learning and artificial intelligence, in general, can do much more and be utilized in things other than specialized disease detection and prediction models. Just as an instance in \cite{mohammadi2020impact, hu2021role, weiss2004social} authors try to analyze the spread of diseases such as influenza and Covid-19 in relation to social behaviors and environmental factors. In this section, we review various tools and fields that benefited from advancements in deep learning and in return made the lives of patients and clinicians a lot easier.

\subsection{\textbf{Internet of Things}} 
As time goes on deep learning gets more and more intertwined with other fields of study and medicine, military, and internet of things (IoT) are no exceptions. IoT is a system of interconnected devices that altogether aims to increase the quality of life of humans and make their lives easier. The sustainable, autonomic, and ubiquitous characteristic of IoT makes it perfect for all sort of tasks such as smart cities \cite{shenavarmasouleh2021embodied, mohammadi2021data}, smart farming \cite{jayaraman2016internet, pivoto2018scientific}, smart homes and grids \cite{komninos2014survey, jiang2018smart}, and smart healthcare \cite{baker2017internet, li2021comprehensive}. Smart healthcare, also known as Ubiquitous health (uHealth), electronic health (eHealth), and mobile health (mHealth), enables us to not only automate a big part of traditional clinical workflow, but also sufficiently monitor patients health in a continuous and ubiquitous manner even after discharge and remotely while at their homes or workplace using modern wearables and implantable devices. Implants are utilized for tasks such as heart pacemakers \cite{difrancesco1993pacemaker}, and glucose monitoring \cite{csoeregi1994design} and are usually inserted into the body by light operations and invasive methods. On the other hand, wearables are worn as accessories. They come in various types such as smartwatches, bracelets, and smart rings and are able to monitor heart rate, blood pressure, sleep time, body temperature, and much more. 

Aside from monitoring features, these devices can also embed a combination of pre-trained complex deep learning models discussed in previous sections in themselves and aid doctors in detecting and predicting various diseases. Also, the collected data would be sent to a central server at fixed intervals or if a significant change is detected via the sensors. Central servers can benefit from larger and more complex data feed and hardware resources to then extract deeper patterns and features and notify the corresponding doctor if needed. Thus, personalized treatments will also be more accessible.

Building on this, various telemonitoring frameworks and systems have been proposed by researchers using different machine learning and deep learning models and diverse transmission network technologies to monitor and solve all sorts of health conditions in varying locations and importances \cite{shaikh2012real, kim2015coexistence, ingole2015implementation}. For instance, \cite{gondalia2018iot} uses all these technologies to monitor the health of war soldiers on the battlefield that altogether accelerating the search and rescue operation if an individual is injured. Aside from this, other living assistants such as smart medicine boxes \cite{li2020secured, kinthada2016emedicare, alex2016modern} have been proposed to track the intake of different medicines and make sure that the patients consume the correct dosage at the right time, and if any deviation occurs they would notify the patient as well as the doctor to address the issue.


\subsection{\textbf{Computer-aided Diagnosis}} 
Many types of imaging and radiology techniques, such as magnetic resonance imaging (MRI), radiography, ultrasound, thermography, and tomography, provide us with detailed and valuable images that are crucial for physicians and researchers and play a pivotal role in healthcare nowadays. Before this, doctors had to do an invasive surgery to perform an autopsy if it was a need to acquire additional info about the patient's condition. But now, imaging techniques offer several ways to prevent unnecessary surgeries and with the help from deep learning, huge multi-modal datasets can be formed using 3D and/or layers of 2D images \cite{feng20192d3d, han2019image}. Datasets containing these images are inherently high-dimensional and cannot be processed using traditional machine learning algorithms in real-time. Thus, deep learning technologies are vital to enable the analysis and visualization of such data and help physicians see and diagnose illnesses more effectively. 

Computer-aided design (CAD) is essential in the development of biomedical systems for a variety of applications. It aids in the detection, diagnosis, prediction, analysis, and categorization of illnesses, as well as the management and delivery of health care.
Thanks to CAD, it is now possible and reasonably simple to use data from medical imaging techniques to construct comprehensive and detailed models of patients, as CAD models are able to capture and represent a patient's unique and complicated organ, bone, and tissue structure. Given these characteristics, aside from the usual use cases of disease diagnosis such as brain tumor \cite{amarapur2020computer}, breast cancer \cite{samala2017multi}, colorectal cancer \cite{ahmad2019artificial}, and lung cancer \cite{thakur2020lung}, they are increasingly being utilized in operations, particularly those involving the implantation of medical devices or prosthetics \cite{lee2017reliability, herschdorfer2021comparison}, as well as creating interactive tools for training future doctors, and surgeons \cite{werz20183d, baillargeon2014living}. Additionally, CAD models are the best approach to track the progression of diseases such as Alzheimer's over time \cite{padilla2011nmf}.

Advancement of deep learning has made it now a part of most of the interdisciplinary research and genomics is not an exception. Gene clustering \cite{thalamuthu2006evaluation, wiwie2015comparing}, gene classification \cite{zeebaree2018gene, mi2013large}, quantitative structure-activity relationships models \cite{ma2015deep}, gene expression \cite{chen2016gene}, and phenotyping \cite{singh2018deep, namin2018deep} are all different areas that genomics benefited from deep learning techniques in recent years as bigger raw sequences of data can be plugged into DL models and yet more complex patterns and relationships can be found faster and more accurately. A deeper level of integration between CAD and genomics gives birth to the computer-aided drug design discipline in which on a molecule-by-molecule level, sophisticated computer modeling of molecular dynamics is employed to anticipate how medications would interact with the biological architecture of the human body. Computational models are used to simulate and occasionally depict atom-to-atom processes. This leads to creating targeted treatment and drug delivery that can target specific cells at the molecular level and as a result, increase the chance of success while considerably reducing the adverse effects of the medication \cite{chen2018rise, gawehn2016deep, liu2021deep}.


\section{\textbf{Challenges}} 

\subsection{\textbf{Interpretability}}
Deep learning and machine learning techniques are wonderful for assessing data, but they might look to humans as black boxes, which is undesirable in healthcare. Because computers are here to assist the healthcare community and allow them to operate side by side with them, the models' outputs and corresponding decisions must be interpretable by physicians and clinicians to further examine them if necessary especially if they're wrong.

Machine learning algorithms have an easier time accomplishing this since their reasoning is typically not overly complex and hence easy to comprehend. For example, decision trees can easily illustrate their "decision tree" to demonstrate all of the underlying factors and features that lead to a certain conclusion. Given the intricate architecture of deep learning models, this would not be a simple operation, and the judgments would be difficult to track. To address this problem, researchers are investigating ways to build rules from neural networks and attempting to identify the most prevalent routes in them in order to make sense of their decisions \cite{wang2018interpret, fu1994rule}. While it comes to processing photos and videos, attention models have also been used to highlight the areas where the model focuses the most when deciding on a topic \cite{wang2017residual}.

\subsection{\textbf{Transfer Learning}}
In healthcare industry incorporating expert knowledge into data is very expensive and time consuming. When the size of the database at hand is small, deep learning models such as CNNs are unable to extract enough features from the data and end up either not learning anything or getting overfit and not being able to perform well on similar unseen data. Transfer learning is the suggested method for overcoming this problem by transferring knowledge from one model to another. The human brain excels in transferring its knowledge from one activity to the next. We seldom learn a task from the scratch, preferring instead to build on previous experience with a related activity or topic and by doing so, we expedite our new learning process. Likewise, When there isn't a good enough dataset for the destination domain, but one exists for the source domain, transfer learning is employed by reusing parts of the model which have been already trained on a similar task as the basis for the new task at hand. This method has been shown to be highly practical. Furthermore, because the pre-trained weights are used, the new model just needs to train the final few layers, and hence it saves a lot of time and computing resources. The remaining layers will be either frozen and left untouched during the training or can be slightly fine-tuned \cite{yosinski2014transferable, weiss2016survey, ying2018transfer}.

\subsection{\textbf{Data Quality}}
Most of the datasets that can be found in the internet do not have exact same number of instances for each of the classes. This can cause a problem in many classification approaches as the model can overfit to the class with the most number of instances and get biased \cite{shenavarmasouleh2019causes} to yield the best accuracy possible while failing to extract useful features. Also, datasets are often noisy and have missing values in them, hence preprocessing techniques should be employed to deal with these issues and normalize the data before training our deep learning model. Additionally, sometimes we encounter the curse of dimensionality and that basically means that we have too many features and thus there's a need to use feature extraction and reduction approaches to remedy this problem \cite{mohammadi2021evolutionary}.
\subsection{\textbf{Interoperability}}
When it comes to IoT, the entire obtained data should be appropriately saved for future use cases. But, this is not as simple as just putting everything into a database. Central servers can exchange data with other services. However, there is no assurance that the other servers will use the same data format standards, thus some kind of standard, such as interfaces or database level rules and schemas, must be incorporated. As a result, when dealing with numerous computer systems sharing and utilizing each other's information, semantic interoperability plays a pivotal role in the system.

Every piece of data must contain meta-information about various entities in order to provide context for the corresponding values to enable the possibility of connecting and linking these small pieces of data, as well as automatic reasoning and inference to eventually transform data into knowledge. The Resource Description Framework (RDF) \cite{miller1998introduction} and the Web Ontology Language (OWL) \cite{mcguinness2004owl, antoniou2004web} are two very distinct but complementary approaches to accomplish these goals, and query languages like SPARQL \cite{perez2009semantics, barbieri2009c} may be used to query data from many sources all across the internet.


\section{\textbf{Conclusion}} 
Machine learning and deep learning have been a big part of every interdiciplinary research and when we talk about big data, health informatics is the first discipline that comes to mind. Given the amount of different DL architectures and the numerous places that each of them can be employed, in this paper, we reviewed the most prominent DL models and went through some of their applications. To conclude, we also debated on some of the challenges that show up when we incorporate DL in healthcare.




\bibliography{bib} 
\bibliographystyle{ieeetr}

\end{document}